\documentclass[11pt,a4paper]{article}
\usepackage[hyperref]{emnlp2020}
\usepackage{times}
\usepackage{latexsym}
\usepackage{epsfig}
\usepackage{graphicx}
\usepackage{amsmath}
\usepackage{amssymb}
\usepackage{bm}
\usepackage{booktabs}
\usepackage{verbatim}
\usepackage{subfigure}
\usepackage{float}

\usepackage{microtype}

\aclfinalcopy 
\def\aclpaperid{608} 

\title{Learning to Stop: A Simple yet Effective Approach\\ to Urban Vision-Language Navigation}

\author{Jiannan Xiang$^\dagger$, Xin Eric Wang$^\ddagger$, William Yang Wang$^\S$ \\
  $^\dagger$University of Science and Technology of China \\
  $^\ddagger$University of California, Santa Cruz \\
  $^\S$University of California, Santa Barbara\\
  \texttt{\small szxjn@mail.ustc.edu.cn, xwang366@ucsc.edu, william@cs.ucsb.edu}
}

\date{}

\begin{document}
\maketitle
\begin{abstract}
Vision-and-Language Navigation (VLN) is a natural language grounding task where an agent learns to follow language instructions and navigate to specified destinations in real-world environments. A key challenge is to recognize and stop at the correct location, especially for complicated outdoor environments. Existing methods treat the STOP action equally as other actions, which results in undesirable behaviors that the agent often fails to stop at the destination even though it might be on the right path.
Therefore, we propose Learning to Stop (\textsc{L2Stop}), a simple yet effective policy module that differentiates STOP and other actions.
Our approach achieves the new state of the art on a challenging urban VLN dataset \textsc{Touchdown}, outperforming the baseline by $6.89\%$ (absolute improvement) on Success weighted by Edit Distance (SED).
\end{abstract}

\section{Introduction}
Vision-and-language navigation (VLN) aims at training an agent to navigate in real environments by following natural language instructions. 
Compared to indoor VLN~\citep{anderson2018vision}, navigation in urban environments~\citep{chen2019touchdown} is particularly challenging, since urban environments are often more diverse and complex. Several research studies \cite{mirowski2018learning,li2019cross,bruce2018learning} have been conducted to solve the problem.
\begin{figure}
\begin{center}
   \includegraphics[width=1\linewidth]{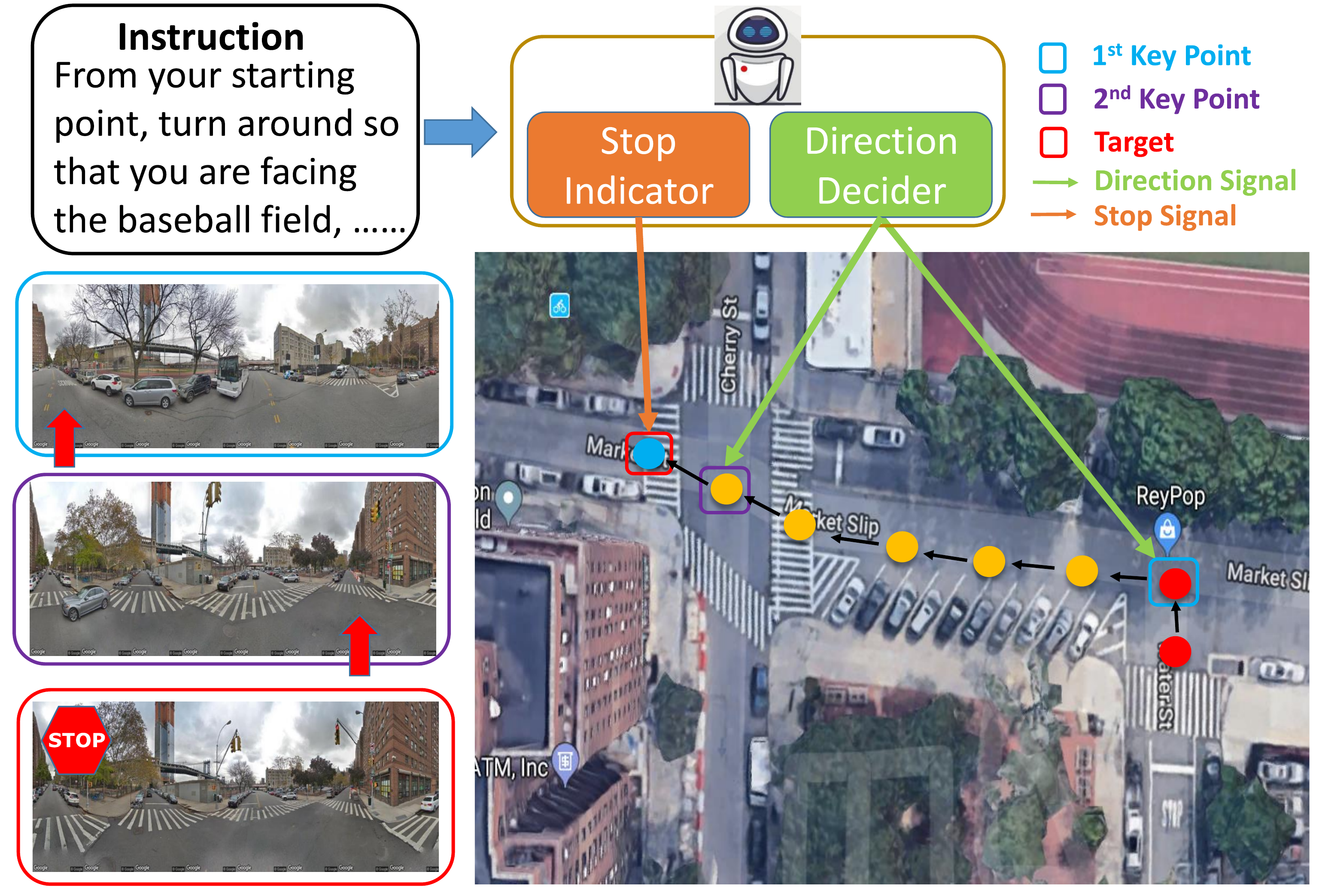}
\end{center}
   \caption{Vision-and-language navigation task in an urban environment. Our \textsc{L2Stop} agent chooses directions at key points and leverages a stop indicator to produce stop or non-stop signals.}
\label{fig:figure1}
\end{figure} 
In this paper, we also focus on the urban VLN task. As shown in Fig.~\ref{fig:figure1}, given a natural language instruction, the agent perceives local visual scene and chooses actions at every time step, learning to match the instruction with the produced trajectory and navigate to the destination. Existing VLN models ~\citep{wang2019reinforced,tan2019learning,ke2019tactical,ma2019regretful,ma2019self, fried2018speaker,wang2018look} seem to neglect the importance of the STOP action and treat all actions equally. 
However, this can lead to undesirable behaviors, also noticed in \citet{cirik2018following, blukis2018mapping}, that the agent fails to stop at the target although it might be on the right path, because the STOP action is severely underestimated.

We argue that the STOP action in the urban VLN tasks is crucially important and deserves special treatment.
First, in contrast to errors on other actions that are likely to be fixed later in the journey, the price of stopping at a wrong location is higher, because producing STOP terminates the episode, and there will be no chance to fix a wrong stop. 
Second, the statistical count of STOP is much lower than other actions as it only appears once per episode. Thus STOP will receive less attention if we treat all actions equally and ignore the difference of occurrence frequency.
Moreover, STOP and other actions need different understandings of the dynamics between the instruction and the visual scene. Both require the alignment between trajectories and instructions, 
but STOP would emphasize the completeness of the instruction and the matching between the inferred target and the surrounding scene, while choosing directions requires a planning ability to imagine the future trajectory.

Therefore, we introduce a Learning to Stop (\textsc{L2Stop}) module to address the issues. \textsc{L2Stop} is a simple and model-agnostic module, which can be easily plugged into VLN models to improve their navigation performance. As we demonstrate in Fig.~\ref{fig:figure1}, the \textsc{L2Stop} module consists of a \textit{Stop Indicator} to determine whether to stop and a \textit{Direction Decider} to choose directions when at key points. Besides, we weigh STOP action more than other actions in the loss function, forcing the agent to pay more attention to the STOP action.
 We conduct experiments on a language-grounded street-view navigation dataset \textsc{Touchdown}\footnote{\url{https://github.com/lil-lab/touchdown}}~\citep{chen2019touchdown}.
Extensive results show that our proposed approach significantly improves the performance over the baseline model on all metrics and achieves the new state-of-the-art on the \textsc{Touchdown} dataset. 
\footnote{The previous version of this work~\cite{xiang2019not} was presented at the NeurIPS 2019 ViGIL workshop.}

\section{Approach}
\label{sec:approach}
Fig.~\ref{fig:figure2} illustrates the framework of our \textsc{L2Stop} model. Specifically, a text encoder and visual encoder are used to get the text and visual representations. Then the trajectory encoder uses the representations to compute the hidden context state, which is the input of the policy module. Unlike previous VLN models, which use one branch policy module, we use our proposed \textsc{L2Stop} module, a two-branch policy module that separates the policies for STOP and other actions. We detail each component below.

\subsection{Visual and Text Encoder}
\begin{figure}
\begin{center}
   \includegraphics[width=1\linewidth]{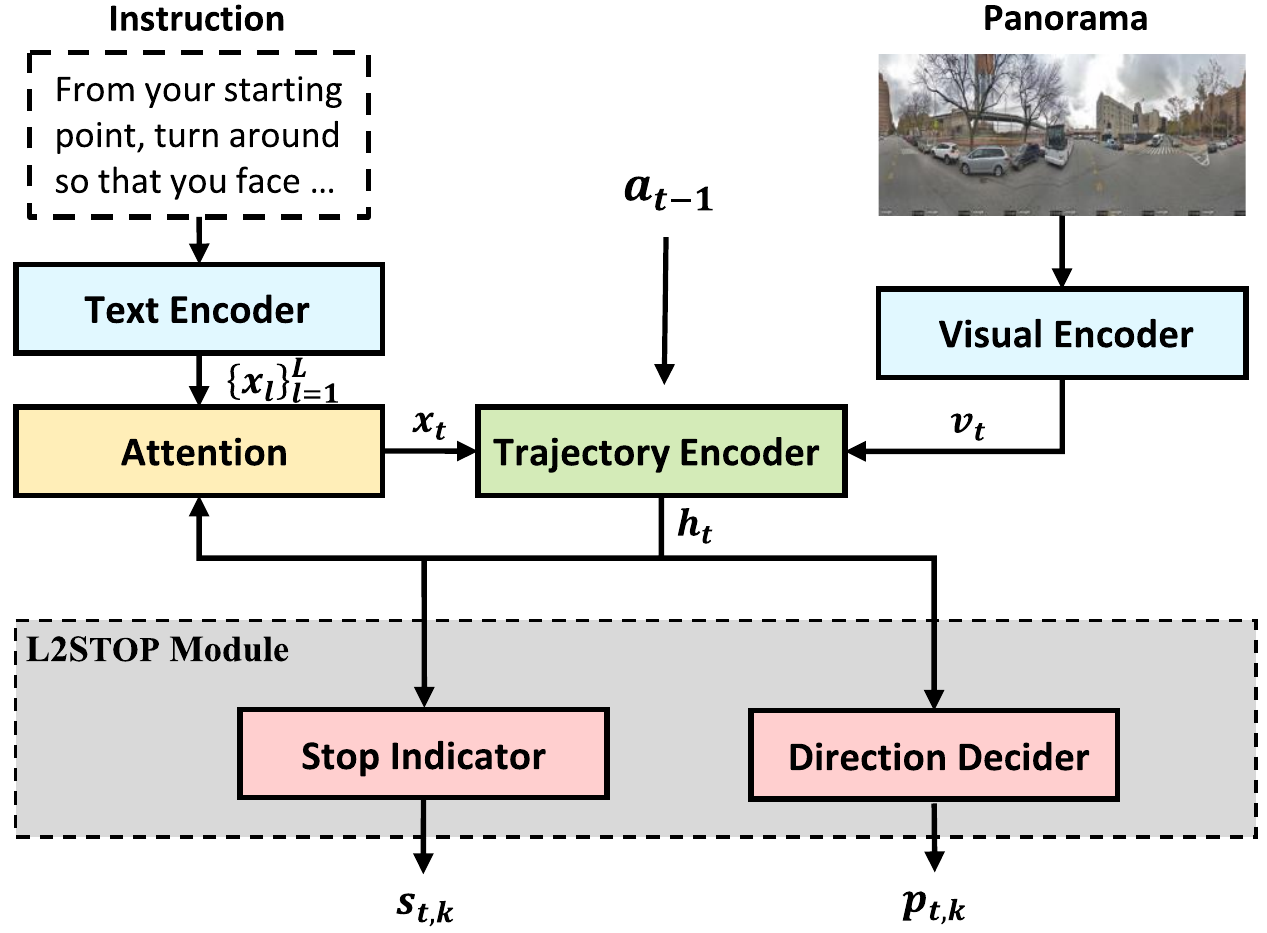}
\end{center}
   \caption{Overview of our \textsc{L2Stop} model.}
\label{fig:figure2}
\end{figure}

As shown in the Fig.~\ref{fig:figure2}, we use two encoders as used in~\citet{chen2019touchdown} for encoding visual scene and language instruction respectively. For visual part, we apply a CNN \citep{krizhevsky2012imagenet} as the visual encoder to extract visual representation $\bm{v}_{t}$ from current visual scene at time step $t$. For text, we adopt an LSTM~\citep{hochreiter1997long} as the text encoder to get the instruction representation $ \bm{X} = \{\bm{x}_1, \bm{x}_2, ..., \bm{x}_l\}$. We then use a soft-attention \citep{vaswani2017attention} to get the grounded textual feature $\bm{x}_t$ at time step $t$:
\begin{align}
    &\alpha_{t, l} = softmax((\bm{W}_x\bm{h}_{t-1})^{\bm{T}}\bm{x}_l)\\
    &\bm{x}_t = \sum\limits_l\alpha_{t, l}\bm{x}_l
\end{align}
where $\bm{W}_x$ denotes parameters to be learnt, $\alpha_{t,l}$ denotes attention weight over $l$-th feature vector at time $t$, and $h_{t-1}$ denotes the hidden context at previous time step.
Then the agent produces the hidden context at the current step: $\bm{h}_t = LSTM([\bm{x}_t, \bm{v}_t, \bm{a}_{t-1}])$.

\begin{table*}[t]
\small
  \centering
  \resizebox{\textwidth}{!}{
  \begin{tabular}{lcccccccccc}
    \toprule
    &\multicolumn{5}{c}{Development}  &\multicolumn{5}{c}{Test}\\
    \cmidrule(r){2-6} \cmidrule(r){7-11}
    Method  &TC$\uparrow$  &SPD$\downarrow$  &SED$\uparrow$  &CLS$\uparrow$  &SDTW$\uparrow$  &TC$\uparrow$  &SPD$\downarrow$  &SED$\uparrow$  &CLS$\uparrow$  &SDTW$\uparrow$\\
    \midrule
    Random  &0.15 &26.63 &0.05 &4.65 &0.06 &0.36 &26.94 &0.01 &4.44 &0.00 \\
    GA      &9.85 &21.43 &9.50 &46.86 &9.44 &9.65 &21.46 &9.21 &47.34 &9.15\\
    \textsc{RConcat} & 11.14 & 19.87 & 10.77 &46.61 &10.76 &9.65 &21.65 &9.45 &44.34 &9.42\\
    \midrule
    \textbf{Ours} &&&&&&&&&&\\ 
    ARC & 15.33 & 18.61 & 14.62 & 48.56 & 14.48 & 14.13 & 19.41 & 13.62 & 48.02 & 13.50\\
    ARC + \textsc{L2Stop} &\textbf{19.48} &\textbf{17.05} &\textbf{19.02}&\textbf{55.68}&\textbf{18.97}&\textbf{16.68}&\textbf{18.84}&\textbf{16.34}&\textbf{53.50}&\textbf{16.34}\\
    \bottomrule
  \end{tabular}}
  \caption{Experimental results on development and test sets.}
  \label{table1}
\end{table*}

\subsection{Learning to Stop Policy Module}
Unlike the existing methods that view all the actions equally important, 
we propose the \textsc{L2Stop} module that helps the agent to learn whether to stop and where to go next with separate policy branches, \emph{Stop Indicator} and \emph{Direction Decider}. 
\paragraph{Stop Indicator}
The stop indicator produces stop or non-stop signals at every time step. At time step $t$, the stop indicator takes the hidden context $\bm{h}_t$ and the time embedding $\bm{t}$ as input and outputs the probabilities of stop and non-stop signals:
\begin{equation}
    s_{t, 1}, s_{t, 2} = softmax(g_2([\bm{h}_t, \bm{t}]))
\end{equation}
where $g_2$ is a linear layer, and $s_{t, 1}$ as well as $s_{t, 2}$ are the probabilities of non-stop and stop signals at time step $t$, respectively. If the stop indicator produces stop signal, the agent will stop immediately. Otherwise, the direction decider will choose a direction to go next.

\paragraph{Direction Decider}
The direction decider is employed to select actions from a subset of the original action space. Specifically, the action subset includes all actions except STOP action (\emph{go forward, turn left, and turn right}).
Empirically, we observe that when navigating in urban environments, the agent only needs to choose directions at the intersections (nodes with more than two neighbors) it encounters in the journey. Therefore, we view these intersections as key points on the road and assume that the direction decider only needs to choose directions at key points and always goes forward otherwise. So at time step $t$, if the agent is at a key point, it will be activated and takes the hidden context $\bm{h}_t$ as well as a learned time embedding $\bm{t}$ as input and outputs the probability of each action in its action space:
\begin{equation}
    p_{t, k} = softmax(g_{1}([\bm{h}_t, \bm{t}]))
\end{equation}
where $g_{1}$ is a linear layer and $p_{t, k}$ is the probability of each action at time step $t$.

\subsection{Learning}
 We use Teacher-Forcing \citep{luong2015effective} method to train the model. We have two loss functions, $\mathcal{L}_{direction}$ and $\mathcal{L}_{stop}$, for direction decider and stop indicator, respectively. $\mathcal{L}_{direction}$ is a regular cross-entropy loss function,
\begin{equation}
    \mathcal{L}_{direction} = -\sum_t\sum_kq_{t,k}log(p_{t,k})
\end{equation}
Where $q_{t,k}$ denotes ground truth label for each action at time step $t$.
For the stop indicator, we use a weighted cross-entropy loss, where we assign a greater weight for the stop signal in the loss function and therefore force the agent to pay more attention to the stop action, in formula,
\begin{equation}
    \mathcal{L}_{stop} = \sum_t - o_{t}log(s_{t, 1})-\lambda (1-o_{t})log(s_{t, 2})\label{Lstop}
\end{equation}
where $o_{t}$ are the ground-truth non-stop signals, and $\lambda$ is the weight for the stop signal. Finally, the agent is optimized with a weighted sum of two loss functions:
\begin{equation}
    \mathcal{L}_{loss} = \gamma \mathcal{L}_{direction} + (1 - \gamma) \mathcal{L}_{stop}\label{Lloss}
\end{equation}
where $\gamma$ is the weight balancing the two losses.

\section{Experiments and Analysis}
\label{experiments}
\subsection{Experimental Settings}
\paragraph{\textsc{Touchdown} Dataset}
We evaluate our approach on the \textsc{Touchdown} dataset \citep{chen2019touchdown} for VLN in real-world urban environment. The navigation environment includes 29,641 panoramas and 61,319 edges from New York City. The dataset contains 9,326 examples of navigation tasks, which are pairs of ground-truth trajectory and instructions describing the trajectory. The dataset is split into training (6,526 examples), development (1,391) and test (1,409) sets.

\paragraph{Evaluation Metrics}
Following \citet{chen2019touchdown}, we report three evaluation metrics for the VLN task in urban environments: Task Completion (TC), Shortest-path Distance (SPD), and Success weighted by Edit Distance (SED). We also add another two metrics evaluating the alignment between the trajectories and the instructions: Coverage weighted by Length Score (CLS)\footnote{\url{https://github.com/google-research/google-research/tree/master/r4r}\label{metrics}} \cite{jain2019stay} and Success weighted by
normalized Dynamic Time Warping (SDTW)\textsuperscript{\ref{metrics}} \cite{magalhaes2019effective}.

\paragraph{Implementation Details} The proposed framework and the baselines are implemented in PyTorch~\cite{paszke2019pytorch}, and the training of the models costs at average 6 hours. We use Adam optimizer~\cite{kingma2014adam} with a learning rate of 0.00025 to train the model. The text encoder consists of a word embedding layer of size 32, and a bi-directional single-layer RNN with 256 hidden units. A single-layer fully connected layer of size 512 is used to map the previous hidden states, which is then used to compute the soft-attention to get the text representation. The visual encoder is a three-layer CNN. The first layer uses 32 $8 \times 8$ kernels with stride 4, and the second layer uses 64 $4 \times 4$ kernels with stride 4, applying ReLu non-linearities after each convolutional operation. Then a single fully-connected layer including biases of size 256 follows. An action embedding layer of size 16 is learned to map the previous action at every time step. Then, we concatenate the text representation, the visual representation, and the action embedding to get the input of the trajectory encoder. The trajectory encoder is a single-layer RNN with 256 hidden states. The time embedding layer is a single fully-connected layer including biases of size 32. Both of the stop indicator and the direction decider consist of a single-layer perceptron with biases and a \textsc{SoftMax} operation to compute the action probabilities.

\begin{table}[t]
\small
  \centering
  \setlength{\tabcolsep}{3pt}
  \resizebox{\linewidth}{!}{
  \begin{tabular}{lccccc}
    \toprule
    Method  &TC$\uparrow$  &SPD$\downarrow$  &SED$\uparrow$  &CLS$\uparrow$  &SDTW$\uparrow$\\
    \midrule
    GA      &9.85 &21.43 &9.50 &46.86 &9.44\\
    GA + \textsc{L2Stop} & 12.58 & 19.76 & 12.18 &50.10 &12.18\\
    \midrule
    \textsc{RConcat} & 11.14 & 19.87 & 10.77 &46.61 &10.76\\
    \textsc{RConcat} + \textsc{L2Stop} &13.01 &19.28 &12.69&50.86&12.66\\
    \bottomrule
  \end{tabular}}
  \caption{Experimental results of the baseline models with and without \textsc{L2Stop} module on the development set.}
  \label{table_}
\end{table}

\subsection{Experimental Results}
We compare the performance of our approach with the baselines: (1) Random: randomly take actions at each time step. (2) GA and \textsc{RConcat}: the baseline models reported in the original dataset paper~\cite{chen2019touchdown}.
We adapt the \textsc{RConcat} model by equipping it with an attention mechanism on instruction representation to get our Attention-RConcat (ARC) model that outperforms \textsc{RConcat}.
Then we integrate ARC with the proposed \textsc{L2Stop} module, which further boosts the performances on all metrics and achieves the best results on both development and test sets.

In Table \ref{table1}, our approach substantially outperforms the baseline models, improving SED from 9.45\% to 16.34\%. Significant improvements on both goal-oriented metrics (TC, SED) and path alignment metrics (CLS, SDTW) demonstrate the effectiveness of \textsc{L2Stop} model in instruction following and goal achievement, which also validate that \textsc{L2Stop} learns not only where to go but also where to stop better.

\begin{table}[t]
\small
  \centering
  \setlength{\tabcolsep}{3pt}
  \resizebox{\linewidth}{!}{
  \begin{tabular}{llccccc}
    \toprule
    \# & Model  &TC$\uparrow$  &SPD$\downarrow$  &SED$\uparrow$  &CLS$\uparrow$  &SDTW$\uparrow$\\
    \midrule
    1 & ARC +\textsc{L2Stop} &\textbf{19.48} &\textbf{17.05} &\textbf{19.02}&\textbf{55.68}&\textbf{18.97}\\
    2 & \quad - one branch &15.40&18.33&14.92&52.00&14.86\\
    3 & \quad - no key points &15.18&18.17&14.55&51.67&14.44\\
    4 & \quad - no weighting &12.65&21.60&12.22&47.91&12.20\\
    \bottomrule
  \end{tabular}}
  \caption{Ablation study results for individual components on the development set.}
  \label{table2}
\end{table}

\subsection{Modularity}
We compare the performance between the baseline models with and without \textsc{L2Stop} module. The results are shown in Table \ref{table_}. Integrated with the \textsc{L2Stop} module, both of the baseline models show improvements on all the metrics. It demonstrates that our approach is model-agnostic and generalizable: the \textsc{L2Stop} module can be plugged into other VLN models and enhance their navigation performance in the urban environment. 

\begin{table}[t]
\small
  \centering
  \begin{tabular}{clc}
    \toprule
    \#&Model&TC (Dev Set)\\
    \midrule
    1&ARC + \textsc{L2Stop}&19.48\\
    2&\quad w/ Oracle Direction &30.63\\
    3&\quad w/ Oracle Stop&61.04\\
    \bottomrule
  \end{tabular}
  \caption{Effect of oracle direction and stop.}
  \label{table:branch}
\end{table}

\begin{figure*}[t]
\begin{center}
   \includegraphics[width=0.8\linewidth]{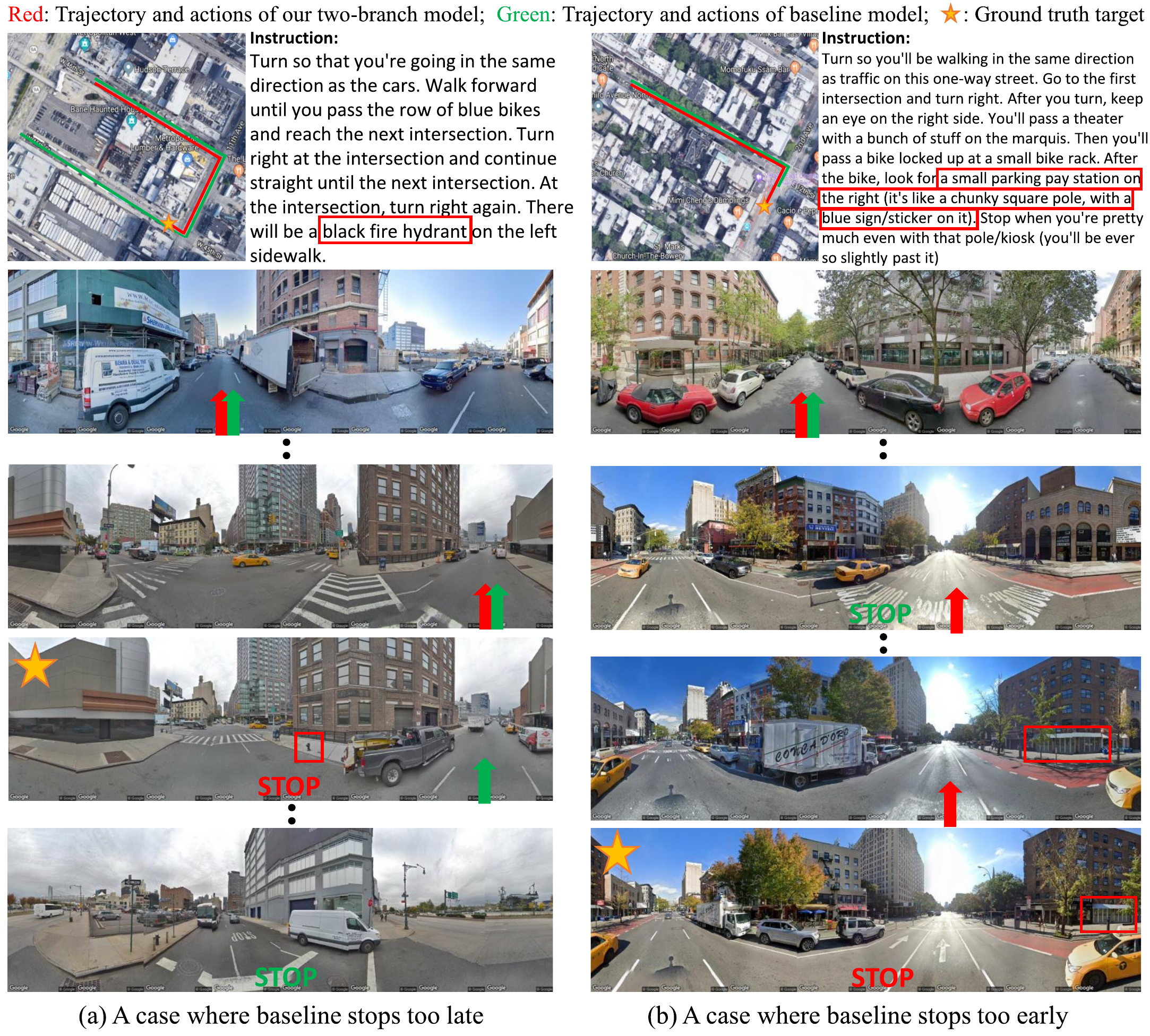}
\end{center}
   \caption{Case study. We choose two cases from the development set, where our proposed model is successful, but the baseline stops either too late or too early. Red boxes show the key items to recognize the target.}
\label{case_study}
\end{figure*}

\subsection{Ablation Study}
\paragraph{Effect of Individual Components}
We conduct an ablation study to illustrate each component's effect on the development set in Table \ref{table2}. 
Row 2-4 shows the influence of each component by removing them respectively from the final model (ARC with \textsc{L2Stop} module). 
Removing any of the components results in worse performance, proving the indispensability of all components in our model.
Row 2 shows the results of ARC with only one policy module, which will disable \emph{turn left} and \emph{ turn right} actions when the agent is not at key points. The results evaluate the effectiveness of the two-branch structure for providing different sub-policies for STOP and other actions. 
Row 3 shows the results of the model whose Direction Decider makes decisions at every time step instead of only at key points. The results validate the effectiveness of only choosing directions at key points. 
Row 4 shows the results where the stop signal's weight is the same as the non-stop signal in the loss function of Stop Indicator. The worst results validate the importance of STOP action. When stop and non-stop signals are treated equally, the agent will prefer non-stop because of its higher occurrence frequency.

\paragraph{Which Is More Important, Stop or Direction?} 
In Table \ref{table:branch}, we study the effect of making either Direction Decider or Stop Indicator an Oracle to see to what extent the model can be improved. 
\textit{Oracle Direction} means the Direction Decider always chooses correct directions, and \textit{Oracle Stop} means the Stop Indicator always produces ground truth stop signals as long as the agent reaches there. \textbf{First}, Row 2 shows the stop branch has about 30\% chance to stop at the right position when the agent is on the right path. \textbf{Second}, The performance in Row 3 is much greater than that in Row 1, indicating that although our approach improves agent's stop ability, the performance is still seriously limited by the wrong stop problem. This indicates that the wrong stop problem in VLN deserves more attention and further study. 

\subsection{Case Study}
We provide visualizations for two qualitative examples to further illustrate how our \textsc{L2Stop} model learns to stop better in Figure~\ref{case_study}. In both cases, our model and the baseline model are on the right path to the target. However, the baseline stops either too late or too early. Specifically, In (a), the baseline agent fails to recognize the black fire hydrant on the target but stops at a place where another black fire hydrant is visible. In (b), the baseline agent successfully recognize the parking pay station on the right, but it ignores the instruction ``slightly past it" and just stops immediately. In contrast, our agent stops in the right place.

\section{Conclusion}
We investigate the importance of the STOP action and study how to learn a policy that can not only make better decisions on where to go but also stop more accurately.   
We propose the \textsc{L2Stop} module for the vision-language navigation task situated in urban environments. 
Experiments illustrate that \textsc{L2Stop} is modular and can be plugged into other VLN models to further boost their performance in urban environments.

\section*{Acknowledgements}
This work was done when the first author was interning at UC Santa Barbara.
The authors would like to thank the anonymous reviewers for the constructive feedback, and Yongji Wu and Yiheng Xu for their helpful discussions.

\bibliography{anthology}
\bibliographystyle{acl_natbib}

\clearpage

\appendix

\section{Appendices}

\subsection{Analysis of Model Structure}

\begin{figure}[htbp]
\begin{center}
   \includegraphics[width=1\linewidth]{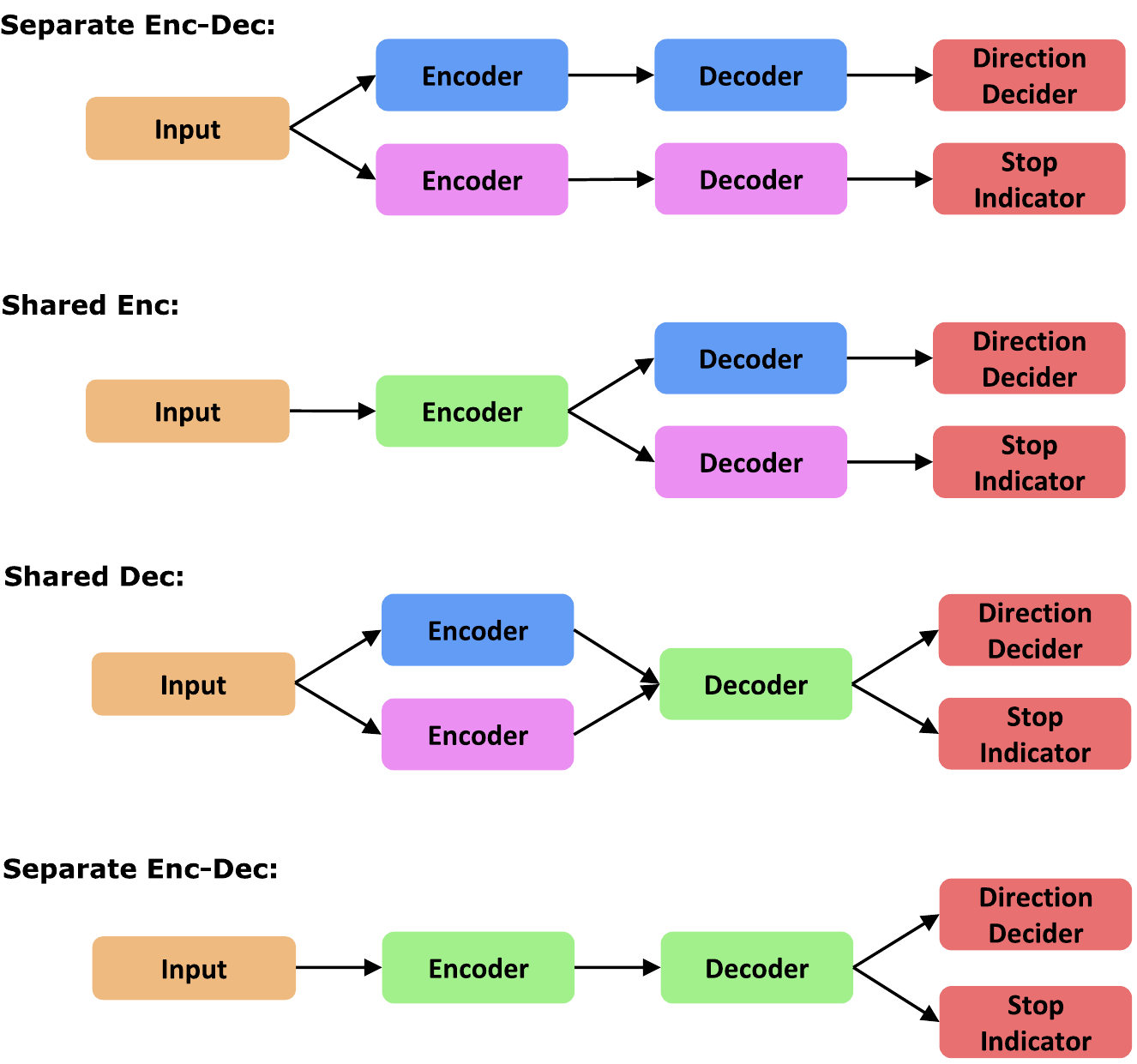}
\end{center}
   \caption{Two-branch VLN models. Input includes language instruction and local visual scene. one Encoder consists of a Visual Encoder and a Text Encoder in Fig.~\ref{fig:figure2}, and Decoder represents Trajectory Encoder in Fig.~\ref{fig:figure2}.}
\label{figure3}
\end{figure}

In Fig.~\ref{figure3}, we examine four model structures to evaluate the interactions between the two branches: (1) \textbf{Separate Enc-Dec} model, where two encoder-decoder models are trained separately for two branches. (2) \textbf{Shared Enc} model, which has a shared encoder but uses two different decoders for two branches. (3) \textbf{Shared Dec} model, which has different encoders for both linguistic and visual input but shared trajectory decoder. (4) \textbf{shared Enc-Dec} model, which shares both the encoder and the decoder. Note that this is the final architecture we use, which is demonstrated in Sec.~\ref{sec:approach}. Table \ref{table:shared} shows the performance of four architectures on the development and test set. \textbf{First}, despite worse performance on other metrics, Separate Enc-Dec can achieve competitive performance on SPD and CLS against other two-branch shared models. The results show that the Separate Enc-Dec agent can produce high-fidelity trajectory matched with instruction but fail to stop at the correct location. This shows that to stop better, the stop indicator requires the information from the direction branch. \textbf{Second}, compared with Shared Enc model, Shared Dec performs competitively on SPD and CLS while much worse on other metrics, indicating that the stop branch learns better from the direction branch in the encoder phase. \textbf{Third}, both Shared Enc and Shared Dec show stronger ability to learn to stop; thus we use Shared Enc-Dec model, which requires fewer parameters. Improved performance shows the Shared Enc-Dec model learns to stop better than other architectures.

\begin{table}[!th]
\small
  \centering
  \setlength{\tabcolsep}{3pt}
  \resizebox{\linewidth}{!}{
  \begin{tabular}{lccccc}
    \toprule
    Method  &TC$\uparrow$  &SPD$\downarrow$  &SED$\uparrow$  &CLS$\uparrow$  &SDTW$\uparrow$\\
    \midrule
    Separate Enc-Dec  &13.71 &\underline{17.67} &13.35 &\underline{55.24} &13.32\\
    Shared Dec & 14.43 & 18.45 & 14.05 & 52.90 & 14.00\\
    Shared Enc & \underline{18.75} & 18.19 & \underline{18.32} &52.42 &\underline{18.27}\\
    Shared Enc-Dec &\textbf{19.48} &\textbf{17.05} &\textbf{19.02}&\textbf{55.68}&\textbf{18.97}\\
    \bottomrule
  \end{tabular}}
  \caption{Performance comparison for four different architectures of the two-branch model on the development set.}
  \label{table:shared}
\end{table}

\subsection{Hyper-Parameters Sensitivity Analysis}
\label{sec44}

\paragraph{Threshold for Stop Signal}

\begin{figure*}[t]
\centering
\subfigure[]{
\begin{minipage}[t]{0.33\linewidth}
\centering
\includegraphics[width=2in]{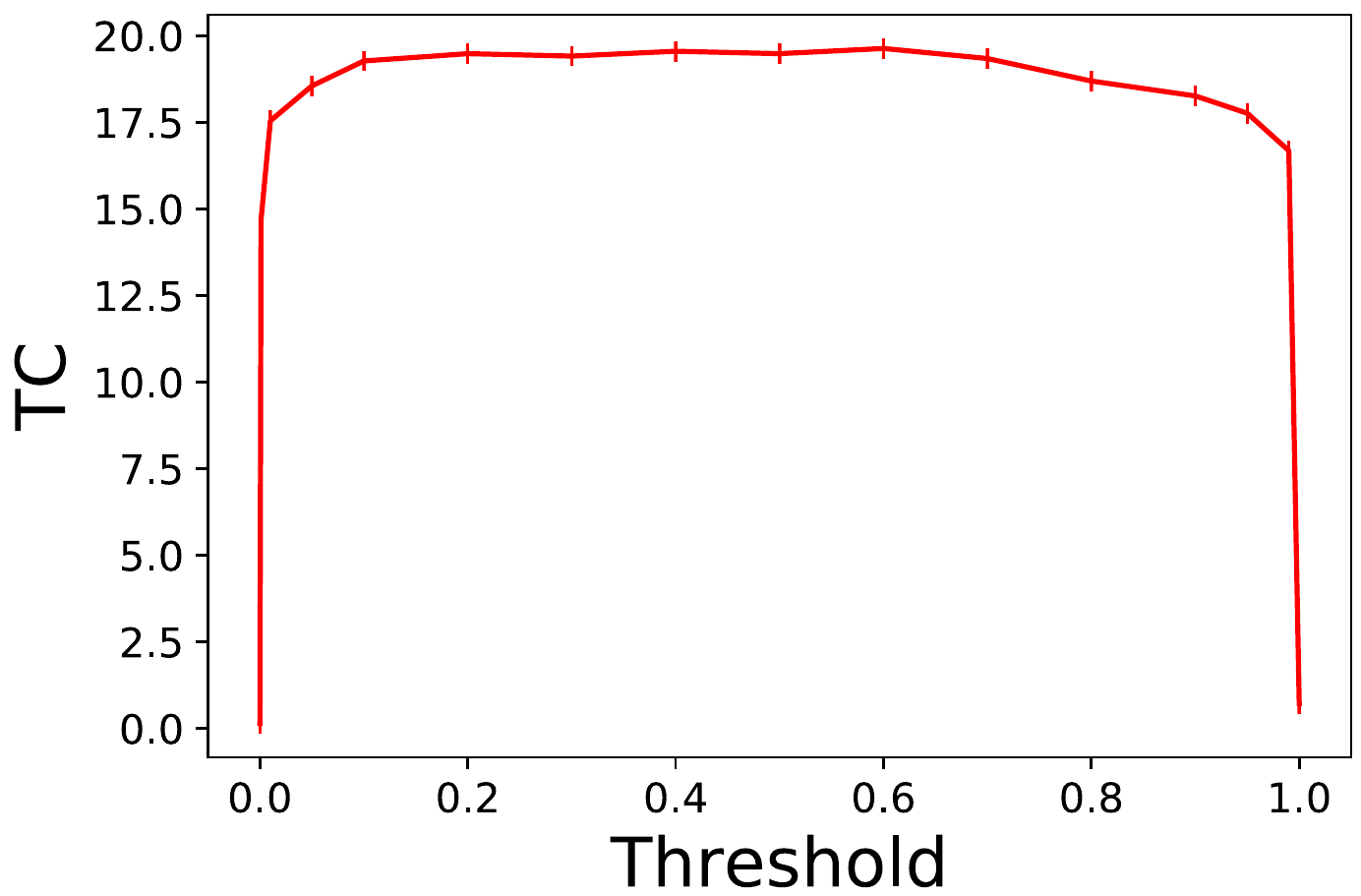}
\end{minipage}%
}%
\subfigure[]{
\begin{minipage}[t]{0.33\linewidth}
\centering
\includegraphics[width=2in]{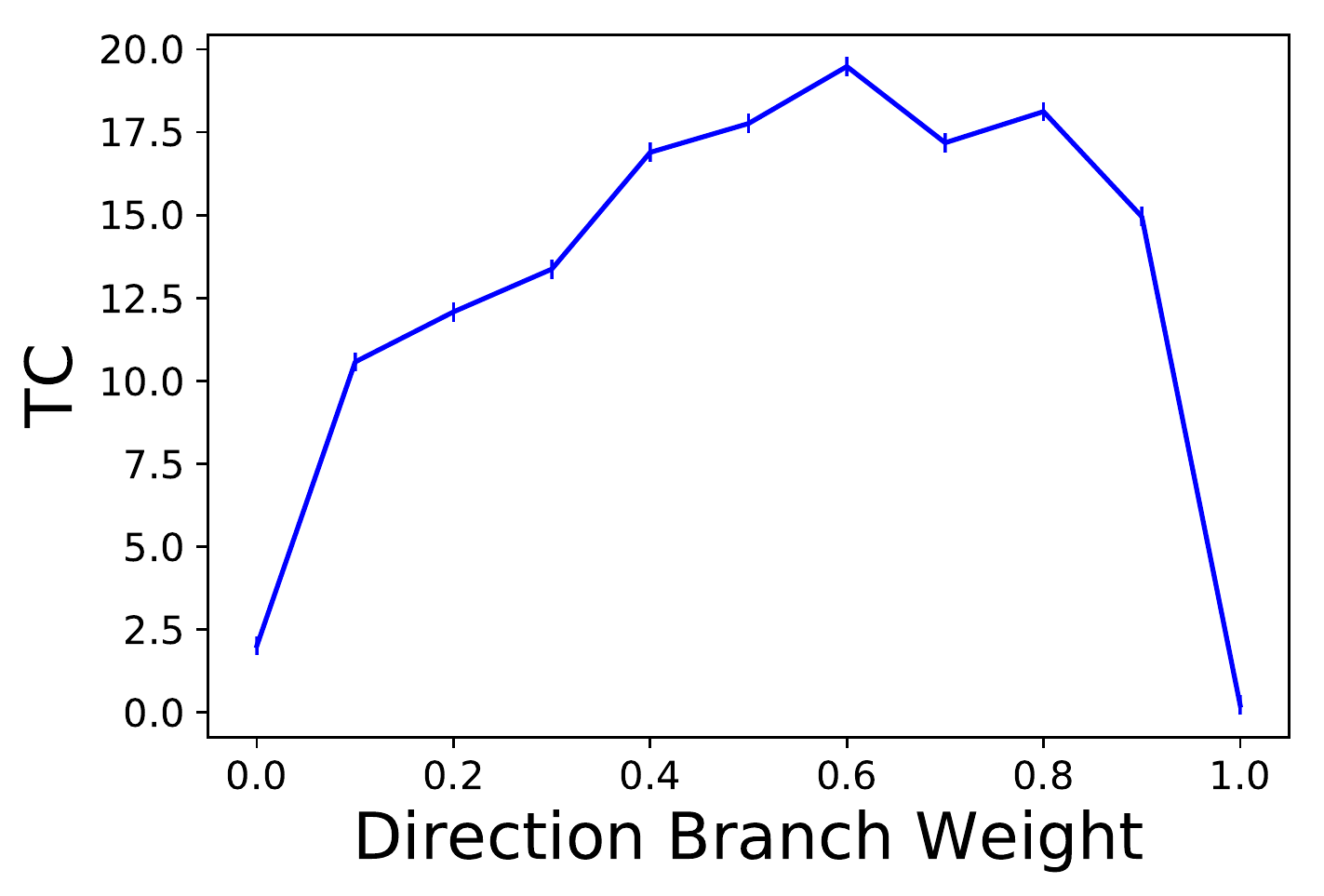}
\end{minipage}%
}%
\subfigure[]{
\begin{minipage}[t]{0.33\linewidth}
\centering
\includegraphics[width=1.95in]{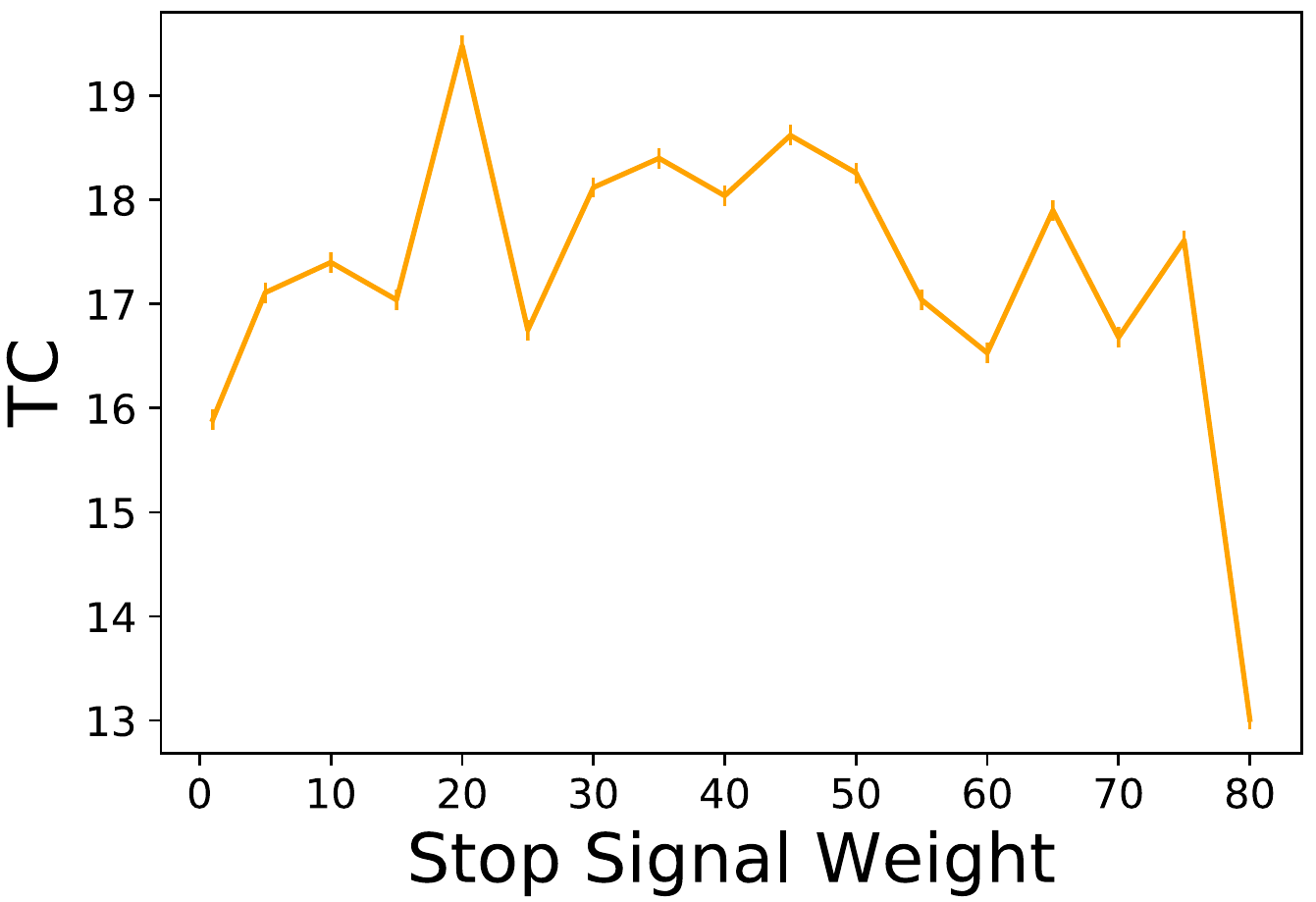}
\end{minipage}
}%
\centering
\caption{(a) Task Completion (TC) scores with different thresholds for the stop signal ($s_{t,2}$ in Equation \ref{Lstop}). TC shows insensitivity to different thresholds. (b) TC scores with different direction branch weights $\gamma$ in Equation \ref{Lloss}. $\gamma=0.6$ gives the highest TC. (c) TC scores with different stop signal weight $\lambda$ in Equation \ref{Lstop}. $\lambda=20$ gives the highest TC. All the experiments are done on the development set,}
\label{figure4}
\end{figure*}

We study the sensitivity of the threshold for stop signals on the development set. The result is shown in Fig.~\ref{figure4} (a). Task-Completion (TC) is consistent in a large range of thresholds, with a slight drop when the threshold is getting higher than 0.7 and sharp decreases when the threshold is close to 0 and 1. The results demonstrate that our approach is insensitive to the change of threshold for stop signals. The consistency of the performance means that the scores of stop signals are either low or high, rarely intermediate. This proves that our approach enables the agent to pay more attention to STOP; that is, the agent is cautious about deciding to stop and only stop when it is highly confident it reaches the goal.

\paragraph{Direction Branch Weight}
We study the sensitivity of direction branch weight $\gamma$ on the development set. The optimal value for $\gamma$ is 0.6, as depicted in Fig.~\ref{figure4} (b), which demonstrates that the balance between the loss functions of two branch enables the agent to not only select correct directions at key points but also stop at the right place. As shown in the figure, smaller $\gamma$ (0-0.5) results in relatively worse performance than higher $\gamma$, indicating that small $\gamma$ enforces the agent to concentrate too much on STOP but ignore the choice for direction. Consistently good performance with larger $\gamma$ (0.6-0.85) shows that only a small weight for the stop branch can significantly improve the agent's stop ability.
\paragraph{Stop Signal Weight}

We study the sensitivity of stop signal weight $\lambda$ on the development set. As shown in the Fig.~\ref{figure4} (c), the optimal value for $\lambda$ is 20. We can see that when $\lambda=0$, our model's performance is similar to the ARC model (15.53 as shown in Table \ref{table1}). However, when setting greater $\lambda$, the TC shows fluctuations, but is consistently better than ARC's performance. Only when $\lambda$ increases to a large number of 80 does the performance decline sharply. This demonstrates the effectiveness of our proposed Weighted Cross-Entropy loss function, which consistently improves the agent's stop ability with a large range of $\lambda$.

\end{document}